\documentclass[11pt,a4paper]{article}
\usepackage[hyperref]{acl2017}
\usepackage{amssymb}
\usepackage{color}
\usepackage{enumitem}
\usepackage{graphicx}
\usepackage{times}
\usepackage{latexsym}
\usepackage{pifont}
\usepackage{url}

\aclfinalcopy 

\newcommand{\unk}{\texttt{unk}}

\title{Stronger Baselines for Trustable Results in Neural Machine Translation}
\author{Michael Denkowski \\
  Amazon.com, Inc. \\
  {\tt mdenkows@amazon.com} \\\And
  Graham Neubig \\
  Carnegie Mellon University \\
  {\tt gneubig@cs.cmu.edu} \\}
\date{}

\begin{document}
\maketitle
\begin{abstract}
  Interest in neural machine translation has grown rapidly as its effectiveness has been demonstrated across language and data scenarios.  New research regularly introduces architectural and algorithmic improvements that lead to significant gains over ``vanilla'' NMT implementations.  However, these new techniques are rarely evaluated in the context of previously published techniques, specifically those that are widely used in state-of-the-art production and shared-task systems.  As a result, it is often difficult to determine whether improvements from research will carry over to systems deployed for real-world use.  In this work, we recommend three specific methods that are relatively easy to implement and result in much stronger experimental systems.  Beyond reporting significantly higher BLEU scores, we conduct an in-depth analysis of where improvements originate and what inherent weaknesses of basic NMT models are being addressed.  We then compare the relative gains afforded by several other techniques proposed in the literature when starting with vanilla systems versus our stronger baselines, showing that experimental conclusions may change depending on the baseline chosen.  This indicates that choosing a strong baseline is crucial for reporting reliable experimental results.
\end{abstract}

\section{Introduction}

In the relatively short time since its introduction, neural machine translation has risen to prominence in both academia and industry.  Neural models have consistently shown top performance in shared evaluation tasks \cite{bojar-EtAl:2016:WMT1,cettoloiwslt} and are becoming the technology of choice for commercial MT service providers \cite{DBLP:journals/corr/WuSCLNMKCGMKSJL16,DBLP:journals/corr/1610.05540}.  New work from the research community regularly introduces model extensions and algorithms that show significant gains over baseline NMT.  However, the continuous improvement of real-world translation systems has led to a substantial performance gap between the first published neural translation models and the current state of the art.  When promising new techniques are only evaluated on very basic NMT systems, it can be difficult to determine how much (if any) improvement will carry over to stronger systems; is new work actually solving new problems or simply re-solving problems that have already been addressed elsewhere?

In this work, we recommend three specific techniques for strengthening NMT systems and empirically demonstrate how their use improves reliability of experimental results.  We analyze in depth how these techniques change the behavior of NMT systems by addressing key weaknesses and discuss how these findings can be used to understand the effect of other types of system extensions.  Our recommended techniques include: (1) a training approach using Adam with multiple restarts and learning rate annealing, (2) sub-word translation via byte pair encoding, and (3) decoding with ensembles of independently trained models.

We begin the paper content by introducing a typical NMT baseline system as our experimental starting point (\S\ref{sec:system}).  We then present and examine the effects of each recommended technique: Adam with multiple restarts and step size annealing (\S\ref{sec:train}), byte pair encoding (\S\ref{sec:subword}), and independent model ensembling (\S\ref{sec:ensemble}).  We show that combining these techniques can lead to a substantial improvement of over 5 BLEU (\S\ref{sec:eval}) and that results for several previously published techniques can dramatically differ (up to being reversed) when evaluated on stronger systems (\S\ref{sec:experiments}).  We then conclude by summarizing our findings (\S\ref{sec:conclusion}).

\section{Experimental Setup}

\subsection{Translation System}
\label{sec:system}

Our starting point for experimentation is a standard baseline neural machine translation system implemented using the Lamtram\footnote{\url{https://github.com/neubig/lamtram}} and DyNet\footnote{\url{https://github.com/clab/dynet}} toolkits \cite{neubig15lamtram,dynet}.  This system uses the attentional encoder-decoder architecture described by \newcite{BahdanauICLR15}, building on work by \newcite{sutskever2014sequence}.  The translation model uses a bi-directional encoder with a single LSTM layer of size 1024, multilayer perceptron attention with a layer size of 1024, and word representations of size 512.  Translation models are trained until perplexity convergence on held-out data using the Adam algorithm with a maximum step size of 0.0002 \cite{kingma2015adam,DBLP:journals/corr/WuSCLNMKCGMKSJL16}.  Maximum training sentence length is set to 100 words.  Model vocabulary is limited to the top 50K source words and 50K target words by frequency, with all others mapped to an \unk{} token.  A post-processing step replaces any \unk{} tokens in system output by attempting a dictionary lookup\footnote{Translation dictionaries are learned from the system's training data using \texttt{fast\_align} \cite{dyer-chahuneau-smith:2013:NAACL-HLT}.} of the corresponding source word (highest attention score) and backing off to copying the source word directly \cite{luong-EtAl:2015:ACL-IJCNLP}.  Experiments in each section evaluate this system against incremental extensions such as improved model vocabulary or training algorithm.  Evaluation is conducted by average BLEU score over multiple independent training runs \cite{papineni-EtAl:2002:ACL,clark-EtAl:2011:ACL-HLT2011}.

\subsection{Data Sets}
\label{sec:data}

We evaluate systems on a selection of public data sets covering a range of data sizes, language directions, and morphological complexities.  These sets, described in Table~\ref{tab:data}, are drawn from shared translation tasks at the 2016 ACL Conference on Machine Translation (WMT16)\footnote{\url{http://statmt.org/wmt16} \cite{bojar-EtAl:2016:WMT1}} and the 2016 International Workshop on Spoken Language Translation (IWSLT16)\footnote{\url{https://workshop2016.iwslt.org}, \url{https://wit3.fbk.eu} \cite{cettoloEtAl:EAMT2012}}.

\begin{table}
  \small\def\arraystretch{1.2}\setlength\tabcolsep{3pt}
  \begin{center}
    \begin{tabular*}{\linewidth}{@{\extracolsep{\fill}}|l|r|l|}
      \hline
      Scenario             & Size (sent) & Sources           \\
      \hline
      WMT German-English   & 4,562,102   & Europarl,         \\
      &             & Common Crawl,     \\
      &             & news commentary   \\
      \hline
      WMT English-Finnish  & 2,079,842   & Europarl,         \\
      &             & Wikipedia titles  \\
      \hline
      WMT Romanian-English & 612,422     & Europarl, SETimes \\

      IWSLT English-French & 220,400     & TED talks         \\

      IWSLT Czech-English  & 114,390     & TED talks         \\
      \hline
    \end{tabular*}
    \vspace{6pt}

    \setlength\tabcolsep{6pt}
    \begin{tabular*}{\linewidth}{@{\extracolsep{\fill}}|l|l|l|}
      \hline
      Scenario & Validation (Dev) Set & Test Set           \\
      \hline
      DE-EN    & News test 2015       & News test 2016     \\

      EN-FI    & News test 2015       & News test 2016     \\

      RO-EN    & News dev 2016        & News test 2016     \\

      EN-FR    & TED test 2013+2014   & TED test 2015+2016 \\

      CS-EN    & TED test 2012+2013   & TED test 2015+2016 \\
      \hline
    \end{tabular*}
  \end{center}
  \caption{\label{tab:data}Top: parallel training data available for all scenarios.  Bottom: validation and test sets.}
\end{table}

\section{Training Algorithms}
\label{sec:train}

\subsection{Background}

The first neural translation models were optimized with stochastic gradient descent \cite{sutskever2014sequence}.  After training for several epochs with a fixed learning rate, the rate is halved at pre-specified intervals.  This widely used rate ``annealing'' technique takes large steps to move parameters from their initial point to a promising part of the search space followed by increasingly smaller steps to explore that part of the space for a good local optimum.  While effective, this approach can be time consuming and relies on hand-crafted learning schedules that may not generalize to different models and data sets.

To eliminate the need for schedules, subsequent NMT work trained models using the Adadelta algorithm, which automatically and continuously adapts learning rates for individual parameters during training \cite{DBLP:journals/corr/abs-1212-5701}.  Model performance is reported to be equivalent to SGD with annealing, though training still takes a considerable amount of time \cite{BahdanauICLR15,sennrich-haddow-birch:2016:P16-12}.  More recent work seeks to accelerate training with the Adam algorithm, which applies momentum on a per-parameter basis and automatically adapts step size subject to a user-specified maximum \cite{kingma2015adam}.  While this can lead to much faster convergence, the resulting models are shown to slightly underperform compared to annealing SGD \cite{DBLP:journals/corr/WuSCLNMKCGMKSJL16}.  However, Adam's speed and reputation of generally being ``good enough'' have made it a popular choice for researchers and NMT toolkit authors\footnote{Adam is the default optimizer for the Lamtram, Nematus (\url{https://github.com/rsennrich/nematus}), and Marian toolkits (\url{https://github.com/amunmt/marian}).} \cite{arthur-neubig-nakamura:2016:EMNLP2016,DBLP:journals/corr/LeeCH16,britz-EtAl:1703.03906,nematus}.

While differences in automatic metric scores between SGD and Adam-trained systems may be relatively small, they raise the more general question of training effectiveness.
In the following section, we explore the relative quality of the optima found by these training algorithms.

\subsection{Results and Analysis}

\begin{figure*}
  \setlength\tabcolsep{4pt}
  \begin{center}
    \begin{minipage}{0.325\textwidth}
      \includegraphics[scale=0.325]{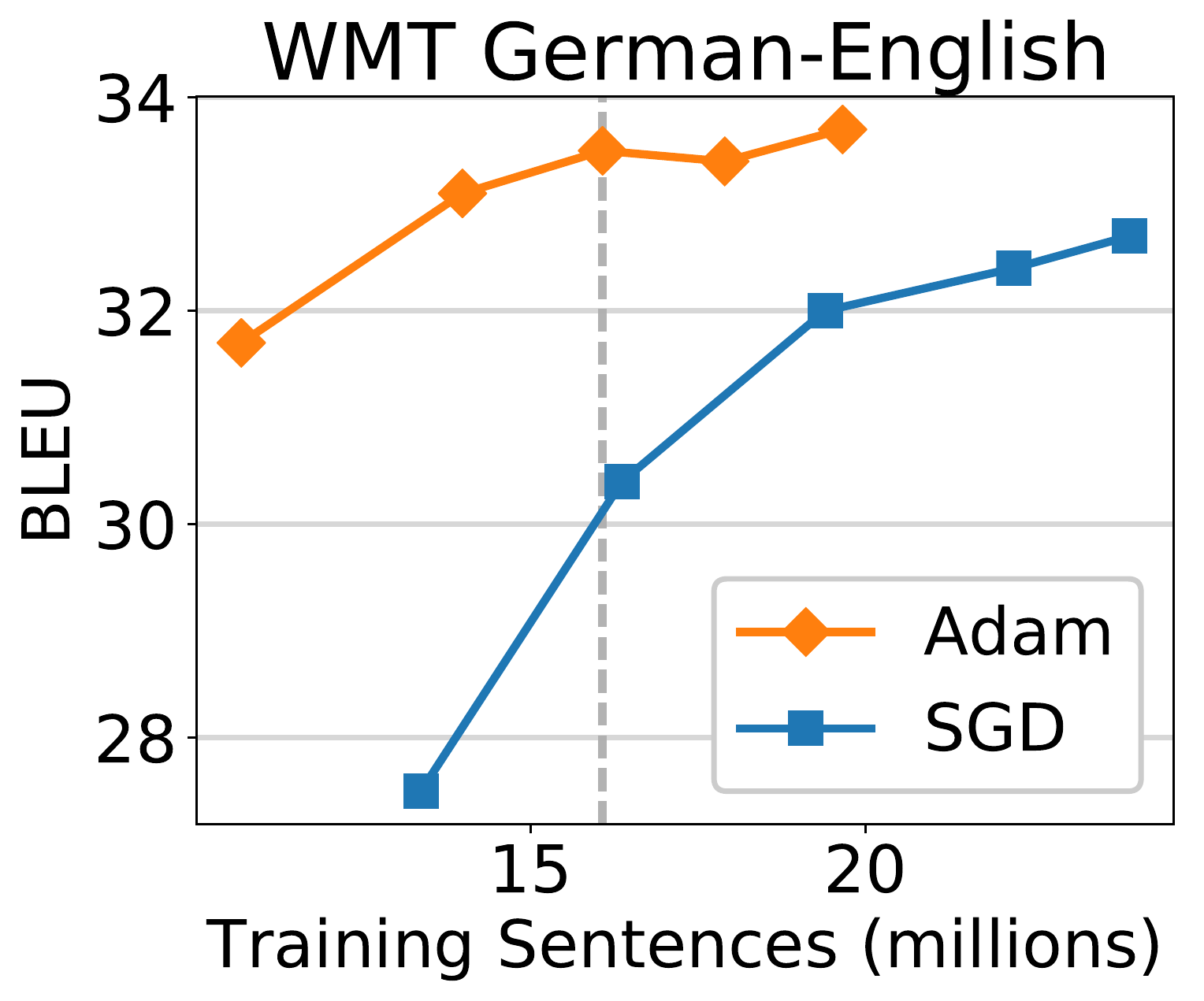}
    \end{minipage}
    \begin{minipage}{0.325\textwidth}
      \includegraphics[scale=0.325]{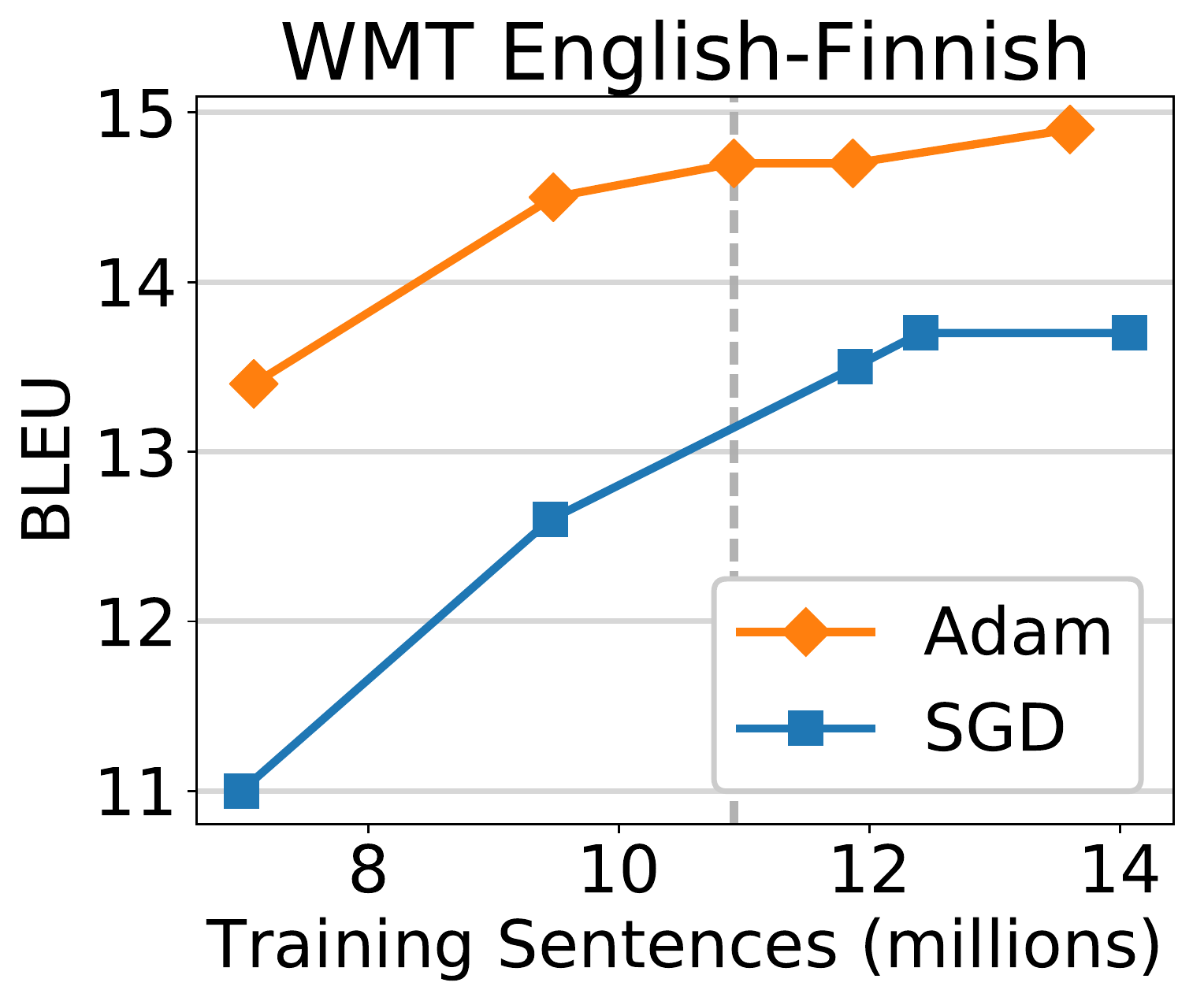}
    \end{minipage}
    \begin{minipage}{0.325\textwidth}
      \includegraphics[scale=0.325]{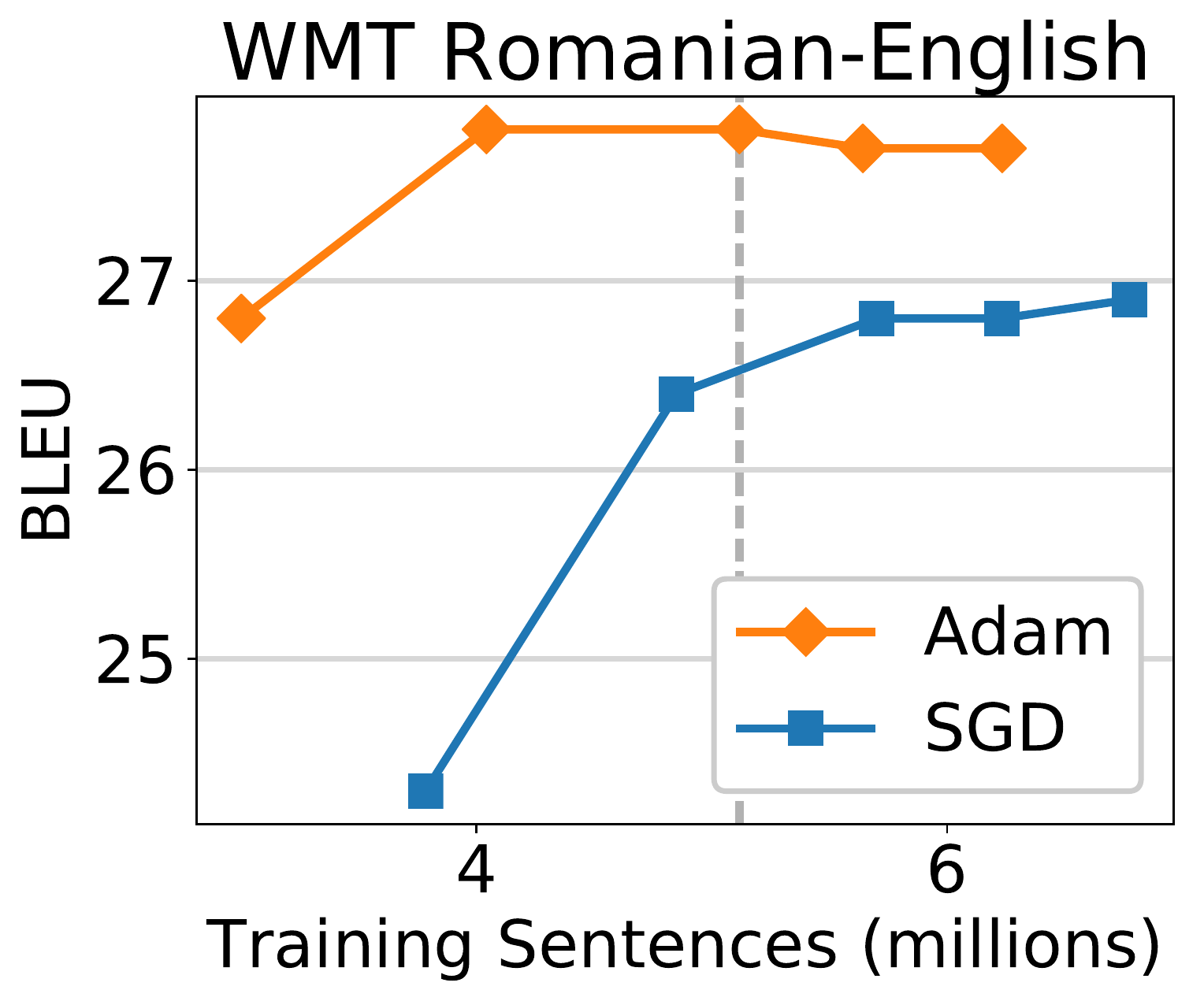}
    \end{minipage}
    \vspace{4pt}

    \begin{minipage}{0.325\textwidth}
      \includegraphics[scale=0.325]{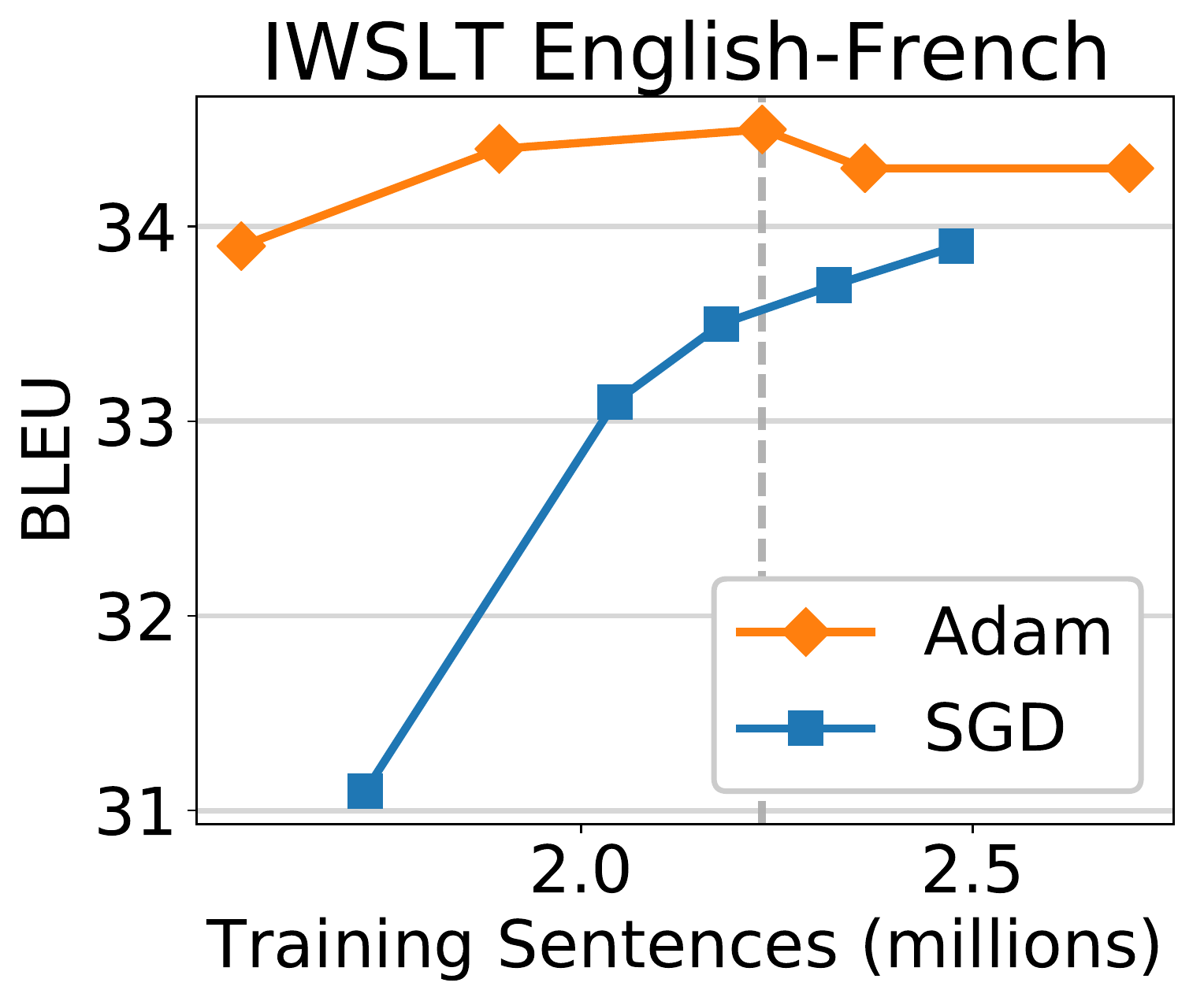}
    \end{minipage}
    \begin{minipage}{0.325\textwidth}
      \includegraphics[scale=0.325]{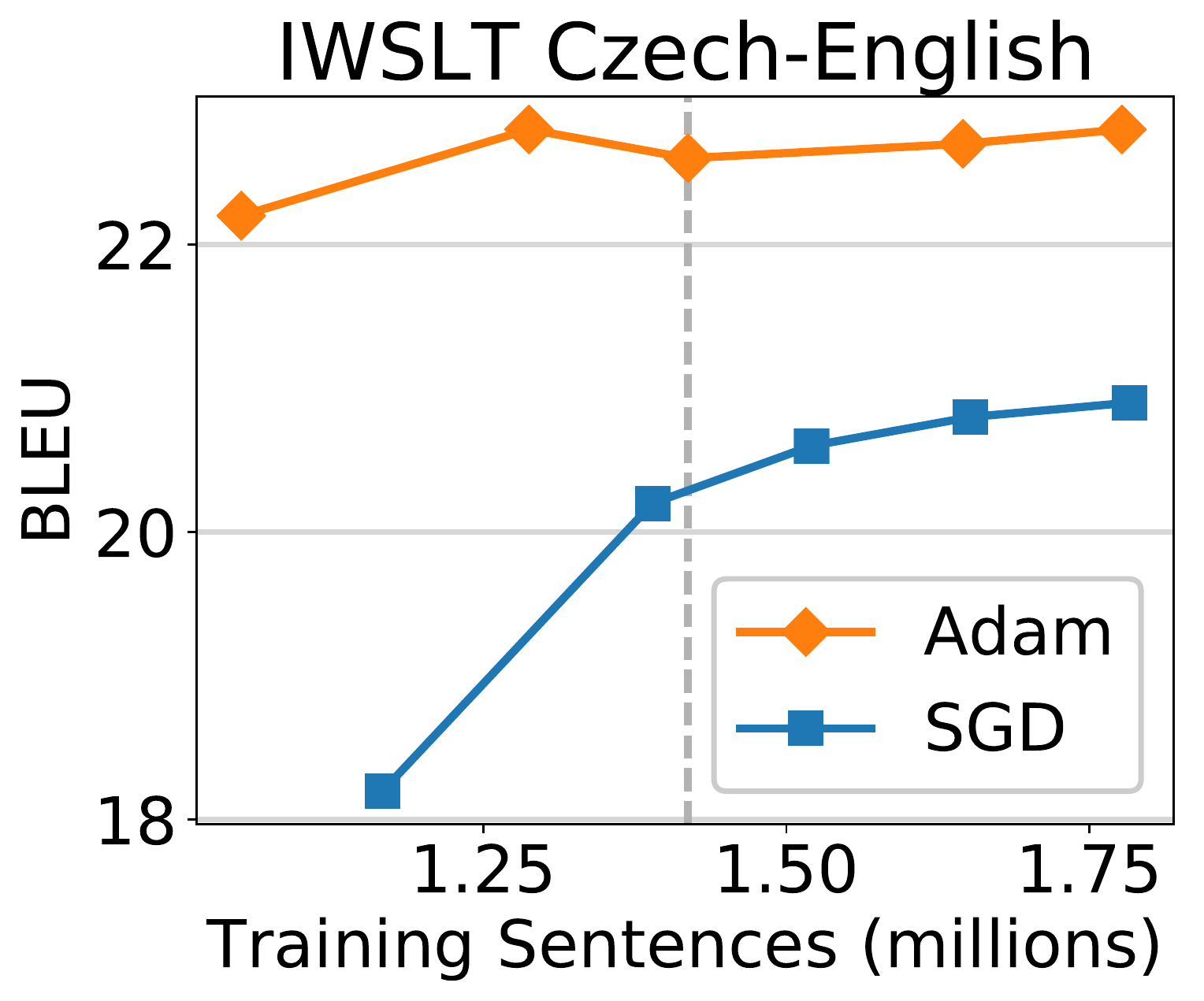}
    \end{minipage}
    \begin{minipage}{0.325\textwidth}
      \begin{center}
        \vspace{7.5pt}
        \scalebox{0.95}{
          \hspace{16pt}\begin{tabular*}{122.5pt}{@{\extracolsep{\fill}}|l|r|r|}
          \hline
          & SGD     & Adam \\
          & (final) & (2r) \\
          \hline
          DE-EN &  6.05 & \textbf{5.86} \\
          EN-FI & 17.97 & \textbf{17.30} \\
          RO-EN & 11.20 & \textbf{10.15} \\
          EN-FR &  6.09 & \textbf{5.85} \\
          CS-EN & 15.16 & \textbf{12.92} \\
          \hline
          \end{tabular*}
        }
        \vspace{4pt}

        \scalebox{0.95}{\hspace{13pt}\small{}Dev Set Perplexities}
      \end{center}
    \end{minipage}
  \end{center}
  \caption{Results of training the NMT models with Adam and SGD using rate annealing.  Each point represents training to convergence with a fixed learning rate and translating the test set.  The learning rate is then halved and training resumed from the previous best point.  Vertical dotted lines indicate 2 Adam restarts.  The table lists dev set perplexities for the final SGD model and the 2-restart Adam model.  All reported values are averaged over 3 independent training runs.}
  \label{fig:optimizers}
\end{figure*}

To compare the behavior of SGD and Adam, we conduct training experiments with all data sets listed in \S\ref{sec:data}.  For each set, we train instances of the baseline model described in \S\ref{sec:system} with both optimizers using empirically effective initial settings.\footnote{Learning rates of 0.5 for SGD and 0.0002 for Adam or very similar are shown to work well in NMT implementations including GNMT \cite{DBLP:journals/corr/WuSCLNMKCGMKSJL16}, Nematus, Marian, and OpenNMT (\url{http://opennmt.net}).}  In the only departure from the described baseline, we use a byte-pair encoded vocabulary with 32K merge operations in place of a limited full-word vocabulary, leading to faster training and higher metric scores (see experiments in \S\ref{sec:subword}).

For SGD, we begin with a learning rate of 0.5 and train the model to convergence as measured by dev set perplexity.  We then halve the learning rate and restart training from the best previous point.  This continues until training has been run a total of 5 times.  The choice of training to convergence is made both to avoid the need for hand-crafted learning schedules and to give the optimizers a better chance to find good neighborhoods to explore.  For Adam, we use a learning rate (maximum step size) of 0.0002.  While Adam's use of momentum can be considered a form of ``self-annealing'', we also evaluate the novel extension of explicitly annealing the maximum step size by applying the same halving and restarting process used for SGD. It is important to note that while restarting SGD has no effect beyond changing the learning rate, restarting Adam causes the optimizer to ``forget'' the per-parameter learning rates and start fresh.

For all training, we use a mini-batch size of 512 words.\footnote{For each mini-batch, sentences are added until the word count is reached.  Counting words versus sentences leads to more uniformly-sized mini-batches.  We choose the size of 512 based on contrastive experiments that found it to be the best balance between speed and effectiveness of updates during training.}  For WMT systems, we evaluate dev set perplexity every 50K training sentences for the first training run and every 25K sentences for subsequent runs.  For IWSLT systems, we evaluate every 25K sentences and then every 6,250 sentences.  Training stops when no improvement in perplexity has been seen in 20 evaluations.  For each experimental condition, we conduct 3 independent optimizer runs and report averaged metric scores.  All training results are visualized in Figure~\ref{fig:optimizers}.

Our first observation is that these experiments are largely in concert with prior work: Adam without annealing (first point) is significantly faster than SGD with annealing (last point) and often comparable or slightly worse in accuracy, with the exception of Czech-English where SGD under-performs.  However, Adam with just 2 restarts and SGD-style rate annealing is actually both faster than the fully annealed SGD and obtains significantly better results in both perplexity and BLEU.  We conjecture that the reason for this is twofold.  First, while Adam has the ability to automatically adjust its learning rate, like SGD it still benefits from an explicit adjustment when it has begun to overfit.  Second, Adam's adaptive learning rates tend to reduce to sub-optimally low values as training progresses, leading to getting stuck in a local optimum.  Restarting training when reducing the learning rate helps jolt the optimizer out of this local optimum and continue to find parameters that are better globally.

\section{Sub-Word Translation}
\label{sec:subword}

\subsection{Background}

Unlike phrase-based approaches, neural translation models must limit source and target vocabulary size to keep computational complexity manageable.  Basic models typically include the most frequent words (30K-50K) plus a single \unk{} token to which all other words are mapped.  As described in \S\ref{sec:system}, \unk{} words generated by the NMT system are translated in post-processing by dictionary lookup or pass-through, often with significantly degraded quality \cite{luong-EtAl:2015:ACL-IJCNLP}.  Real-world NMT systems frequently sidestep this problem with sub-word translation, where models operate on a fixed number of word pieces that can be chained together to form words in an arbitrarily large vocabulary.  In this section, we examine the impact of sub-words on NMT, specifically when using the technique of \textit{byte pair encoding} \cite{sennrich-haddow-birch:2016:P16-12}.  Given the full parallel corpus (concatenation of source and target sides), BPE first splits all words into individual characters and then begins merging the most frequently adjacent pairs.  Merged pairs become single units that are candidates for further merging and the process continues to build larger word pieces for a fixed number of operations.  The final result is an encoded corpus where the most frequent words are single pieces and less frequent words are split into multiple, higher frequency pieces.  At test time, words are split using the operations learned during training, allowing the model to translate with a nearly open vocabulary.\footnote{It is possible that certain intermediate word pieces will not appear in the encoded training data (and thus the model's vocabulary) if all occurrences are merged into larger units.  If these pieces appear in test data and are not merged, they will be true OOVs for the model.  For this reason, we map singleton word pieces in the training data to \unk{} so the model has some ability to handle these cases (dictionary lookup or pass-through).}  The model vocabulary size grows with and is limited by the number of merge operations.  While prior work has focused on using sub-words as a method for translating unseen words in morphologically rich languages \cite{sennrich-haddow-birch:2016:P16-12} or reducing model size \cite{DBLP:journals/corr/WuSCLNMKCGMKSJL16}, we examine how using BPE actually leads to broad improvement by addressing inherent weaknesses of word-level NMT.

\subsection{Results and Analysis}

\begin{table}
  \small\def\arraystretch{1.2}\setlength\tabcolsep{4pt}
  \begin{center}
    \begin{tabular*}{\linewidth}{@{\extracolsep{\fill}}|l|ccc|cc|}
      \hline
      & \multicolumn{3}{c|}{WMT} & \multicolumn{2}{c|}{IWSLT} \\
      & DE-EN         & EN-FI         & RO-EN         & EN-FR         & CS-EN         \\
      \hline
      Words 50K & 31.6          & 12.6          & 27.1          & 33.6          & 21.0          \\
      BPE 32K   & \textbf{33.5} & \textbf{14.7} & \textbf{27.8} & 34.5          & 22.6          \\
      BPE 16K   & 33.1          & \textbf{14.7} & \textbf{27.8} & \textbf{34.8} & \textbf{23.0} \\
      \hline
    \end{tabular*}
  \end{center}
  \caption{\label{tab:vocab}BLEU scores for training NMT models with full word and byte pair encoded vocabularies.  Full word models limit vocabulary size to 50K.  All models are trained with annealing Adam and scores are averaged over 3 optimizer runs.}
\end{table}

We measure the effects of byte pair encoding by training full-word and BPE systems for all data sets as described in \S\ref{sec:system} with the incremental improvement of using Adam with rate annealing (\S\ref{sec:train}).  As \newcite{DBLP:journals/corr/WuSCLNMKCGMKSJL16} show different levels of effectiveness for different sub-word vocabulary sizes, we evaluate running BPE with 16K and 32K merge operations.  As shown in Table~\ref{tab:vocab}, sub-word systems outperform full-word systems across the board, despite having fewer total parameters.  Systems built on larger data generally benefit from larger vocabularies while smaller systems perform better with smaller vocabularies.  Based on these results, we recommend 32K as a generally effective vocabulary size and 16K as a contrastive condition when building systems on less than 1 million parallel sentences.

\begin{figure*}
  \begin{minipage}{0.67\textwidth}
    \setlength\tabcolsep{1 pt}
    \begin{tabular}{cc}
      \raisebox{9.5pt}{\includegraphics[scale=0.325]{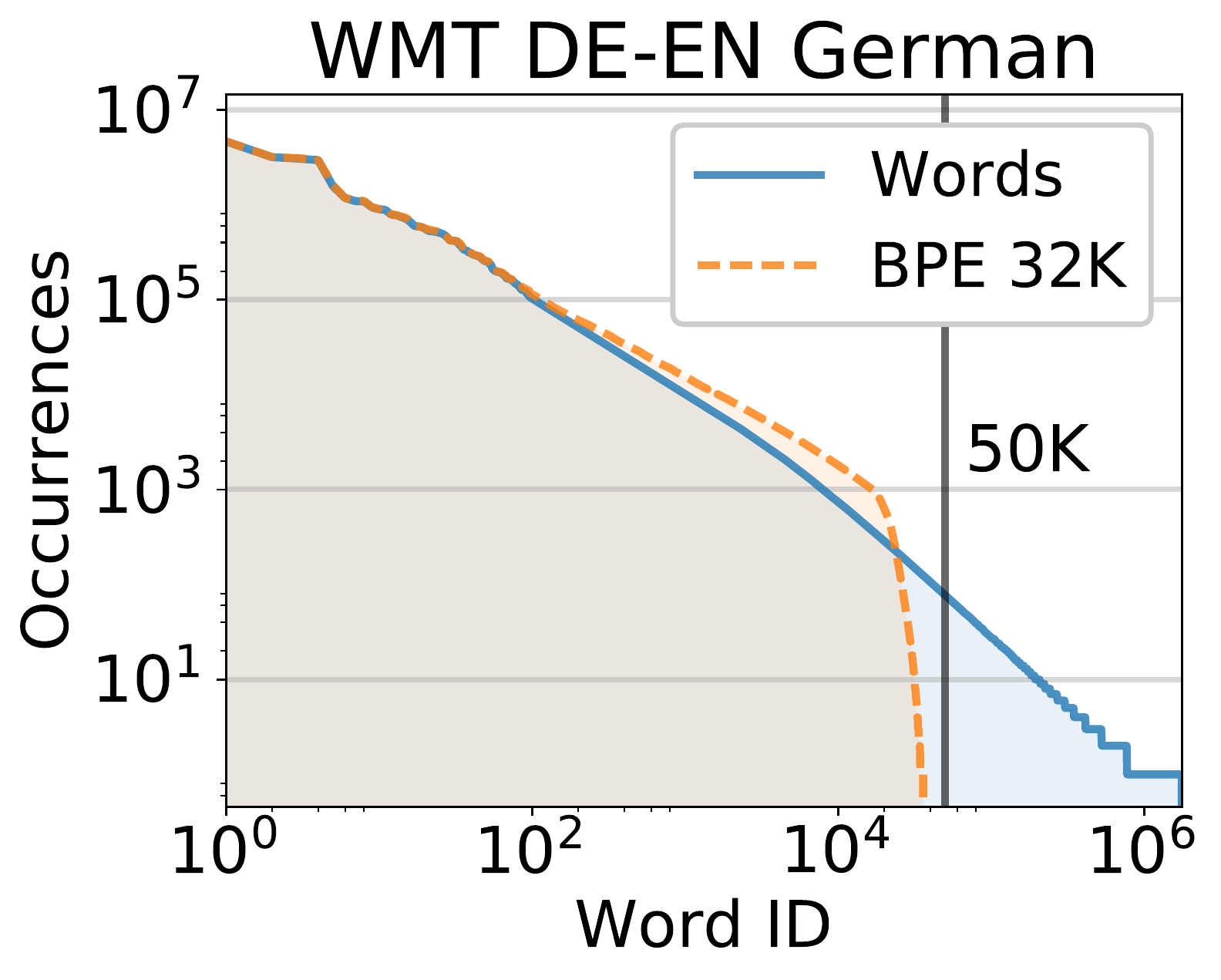}}   & \includegraphics[scale=0.325]{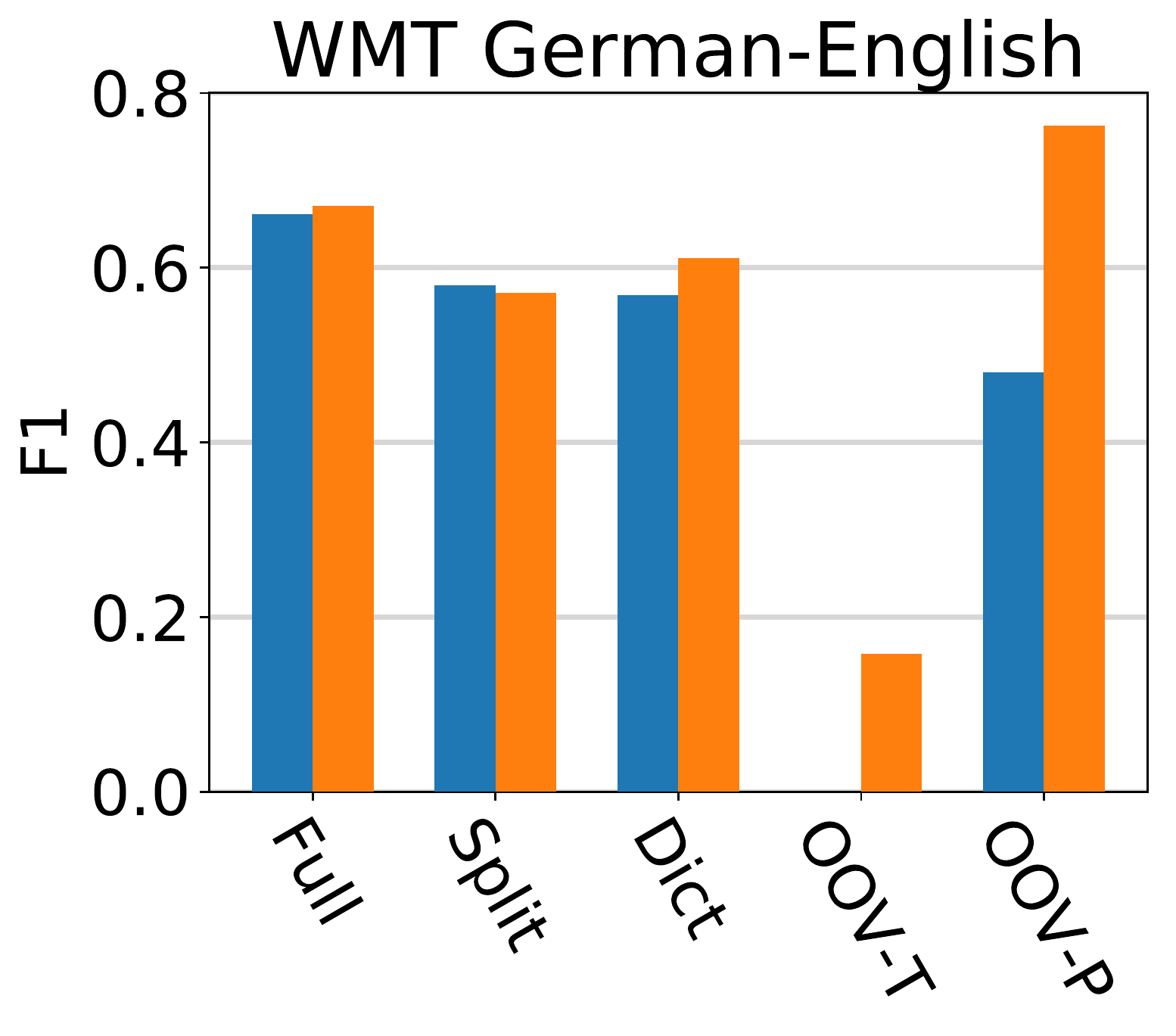}   \\
      \raisebox{9.5pt}{\includegraphics[scale=0.325]{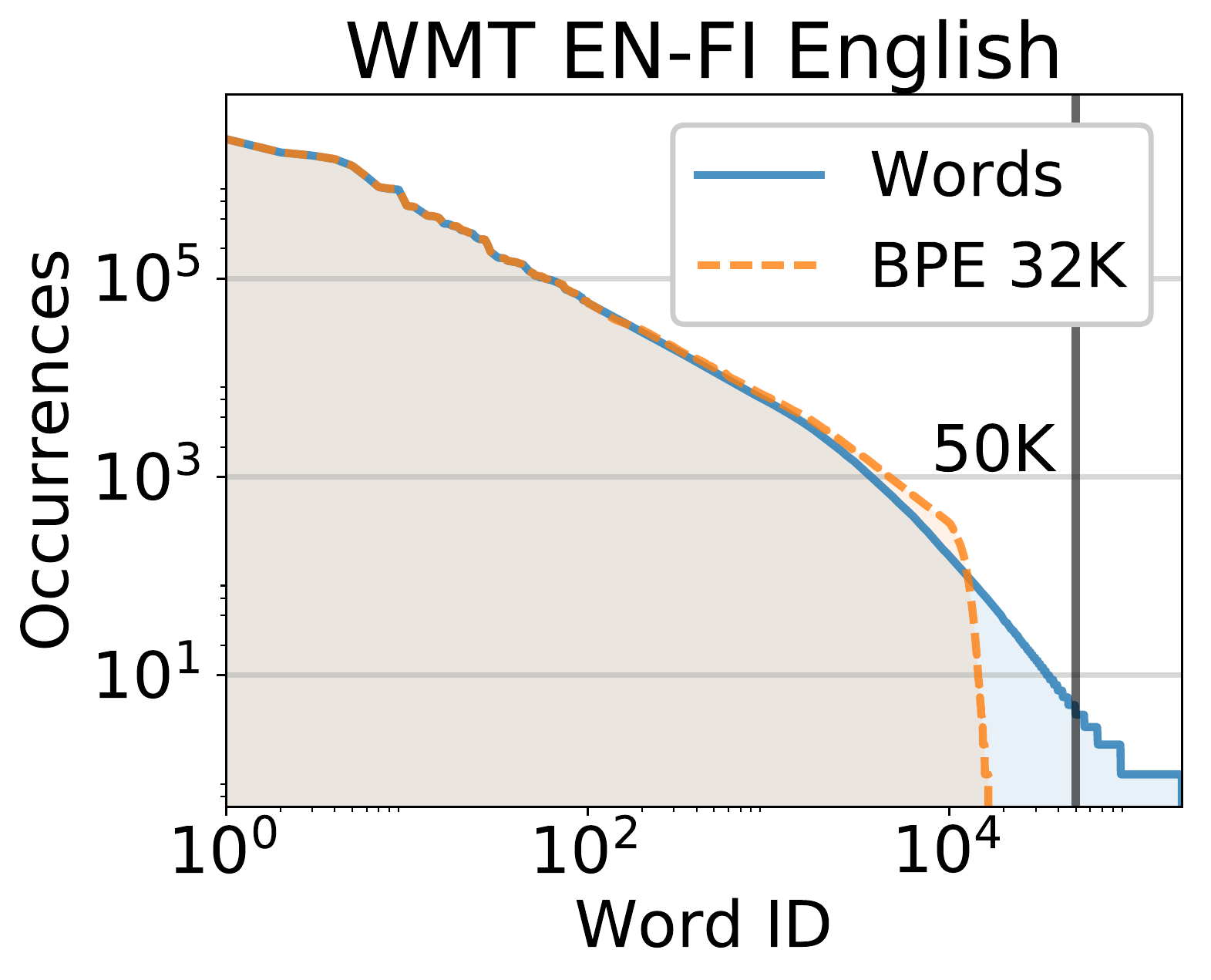}}   & \includegraphics[scale=0.325]{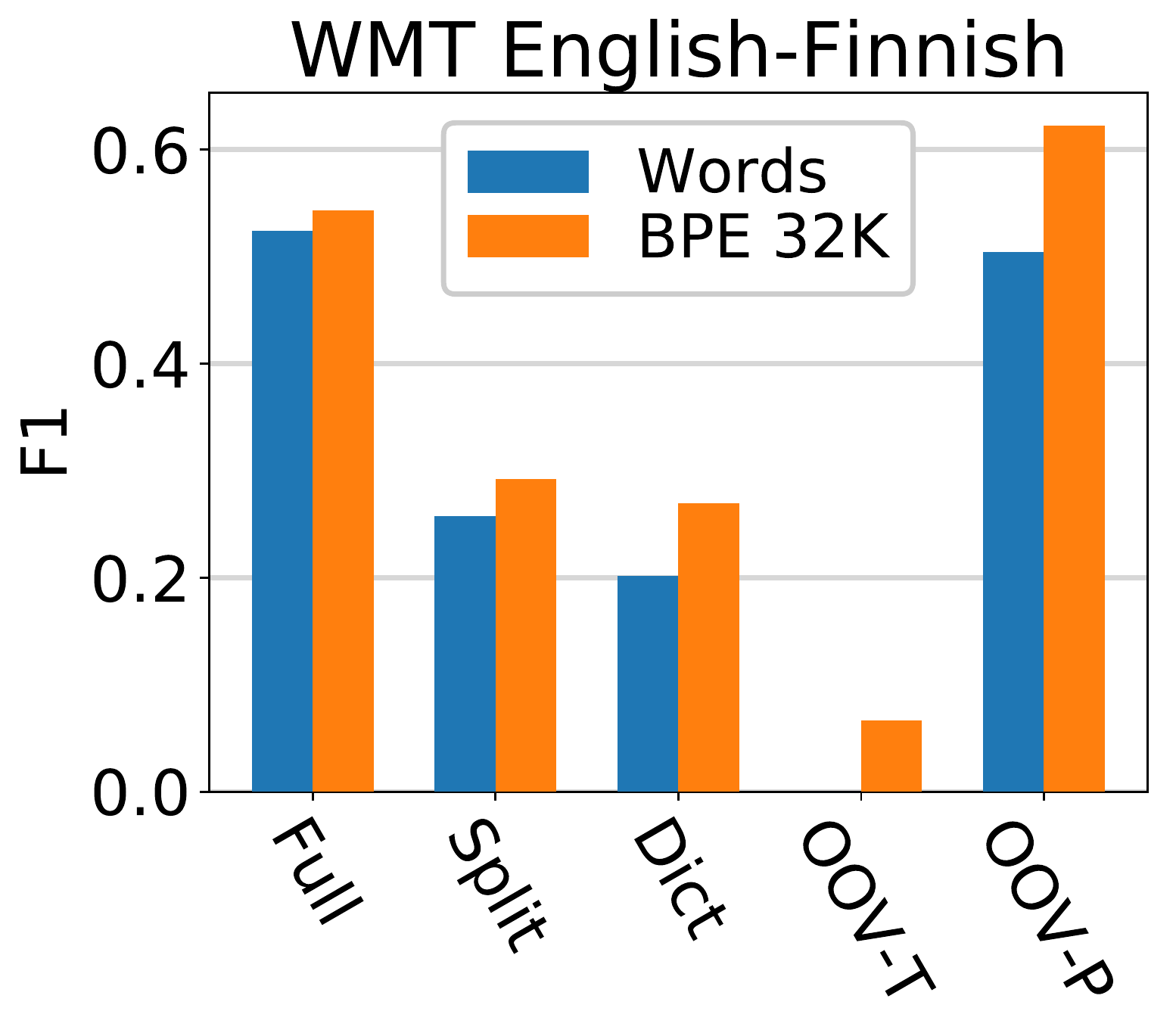}   \\
      \raisebox{9.5pt}{\includegraphics[scale=0.325]{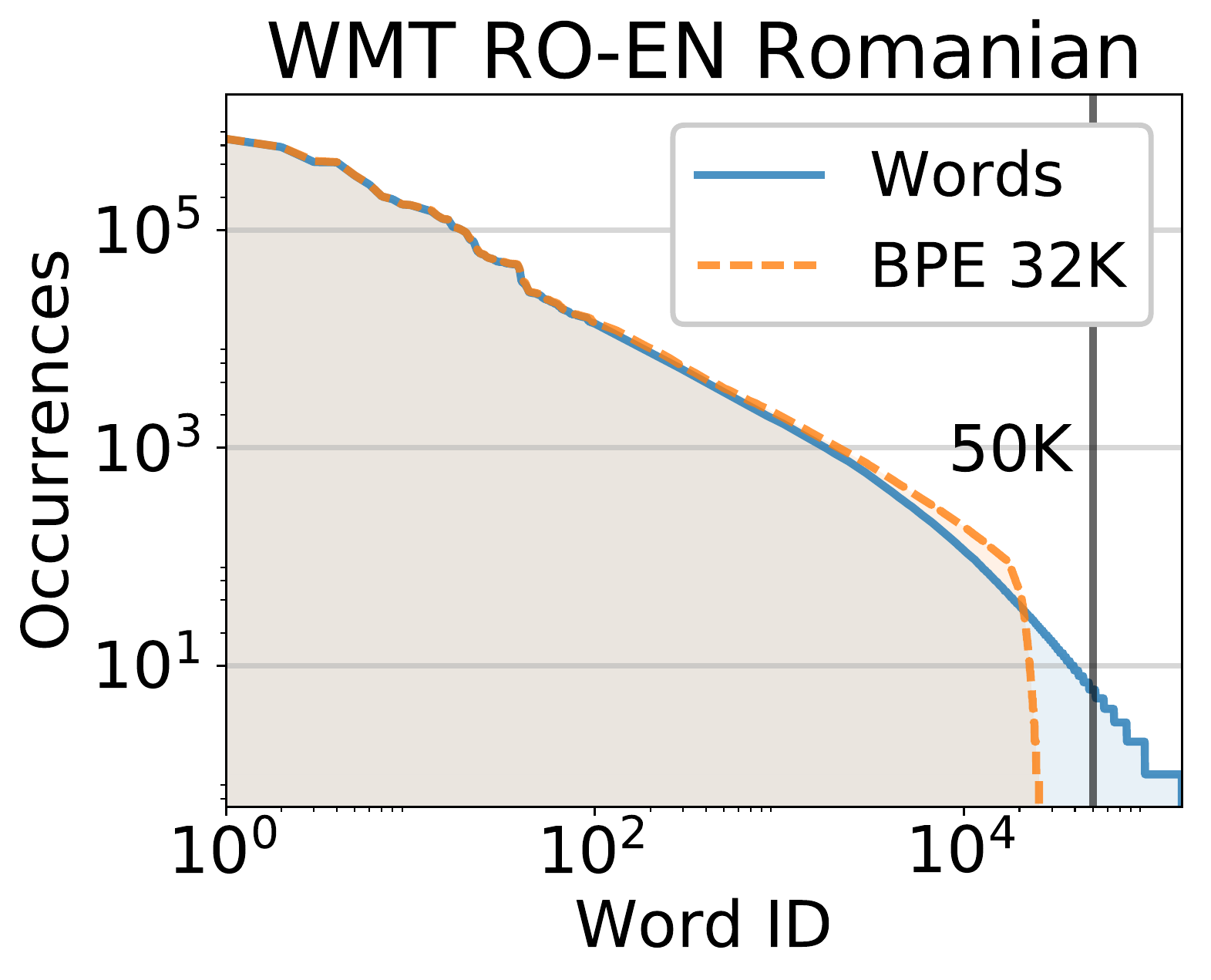}}   & \includegraphics[scale=0.325]{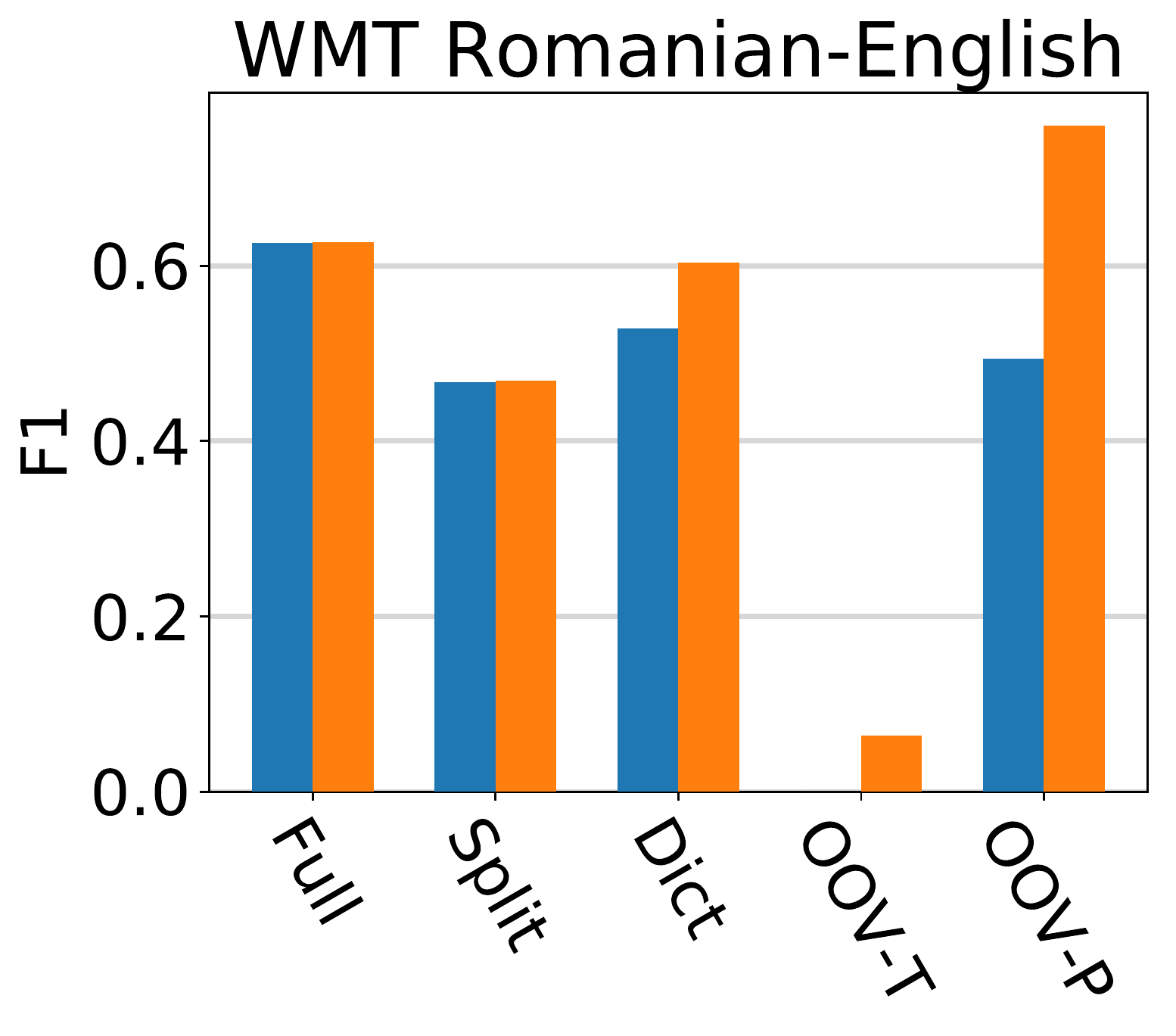}   \\
      \raisebox{9.5pt}{\includegraphics[scale=0.325]{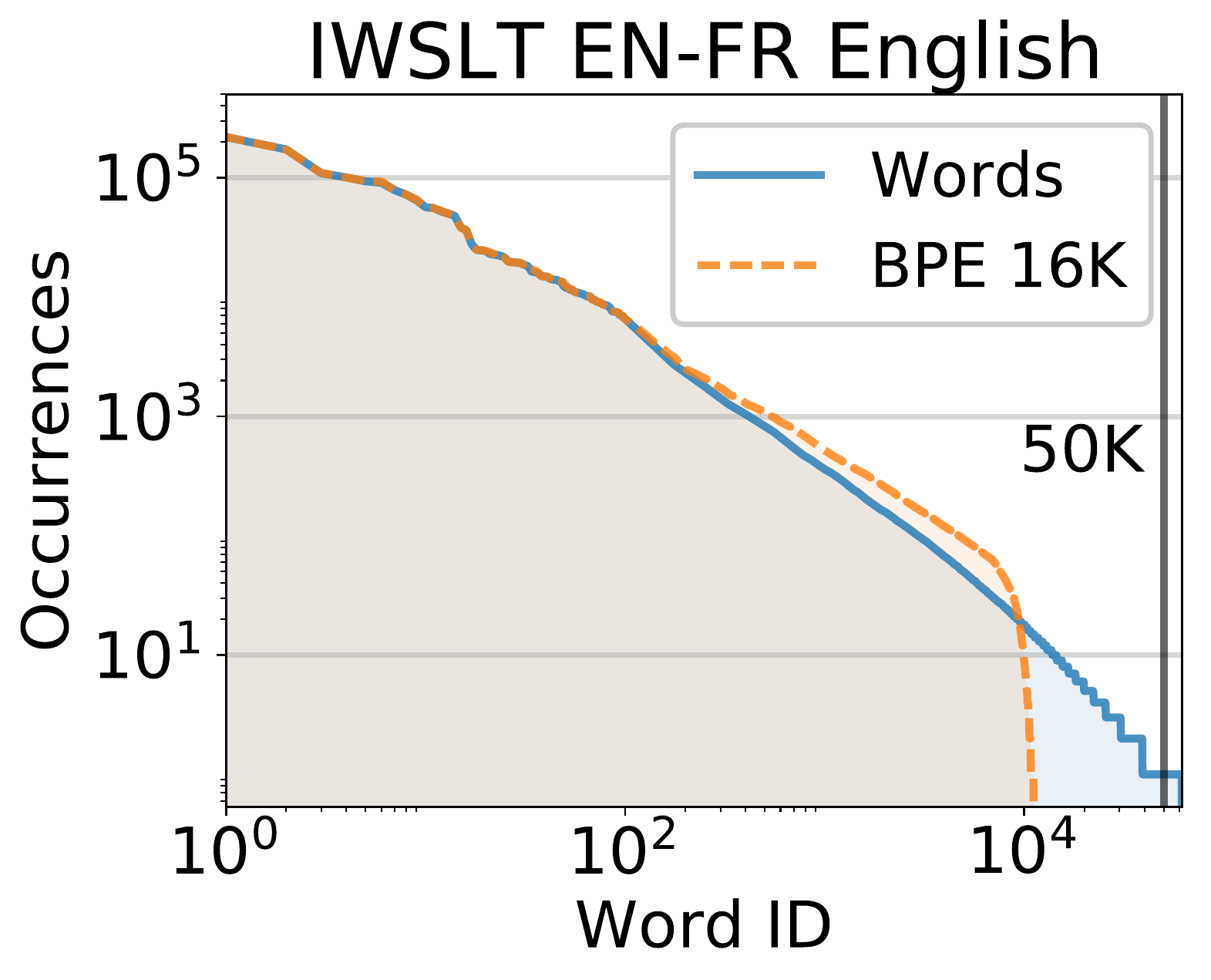}} & \includegraphics[scale=0.325]{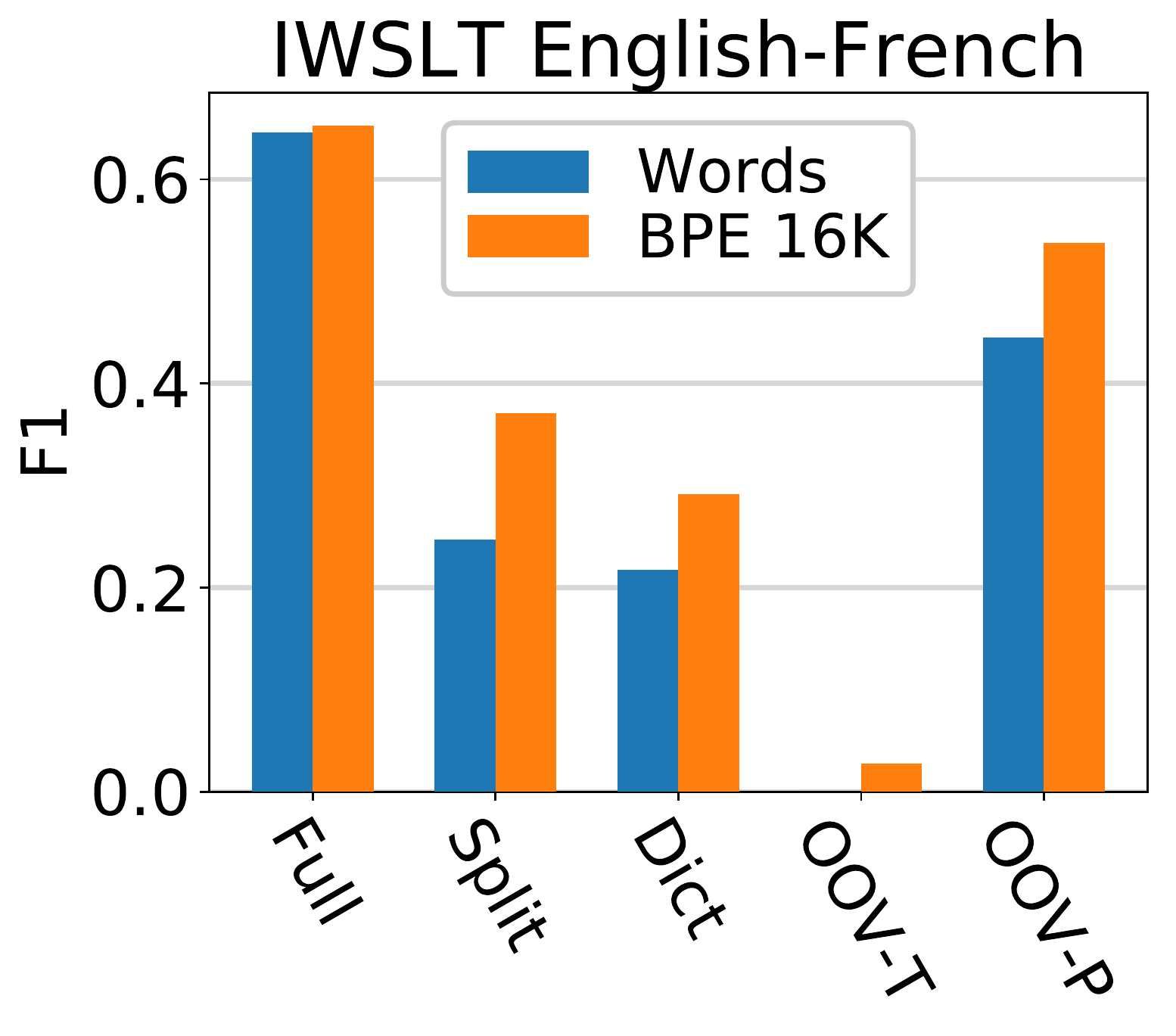} \\
      \raisebox{9.5pt}{\includegraphics[scale=0.325]{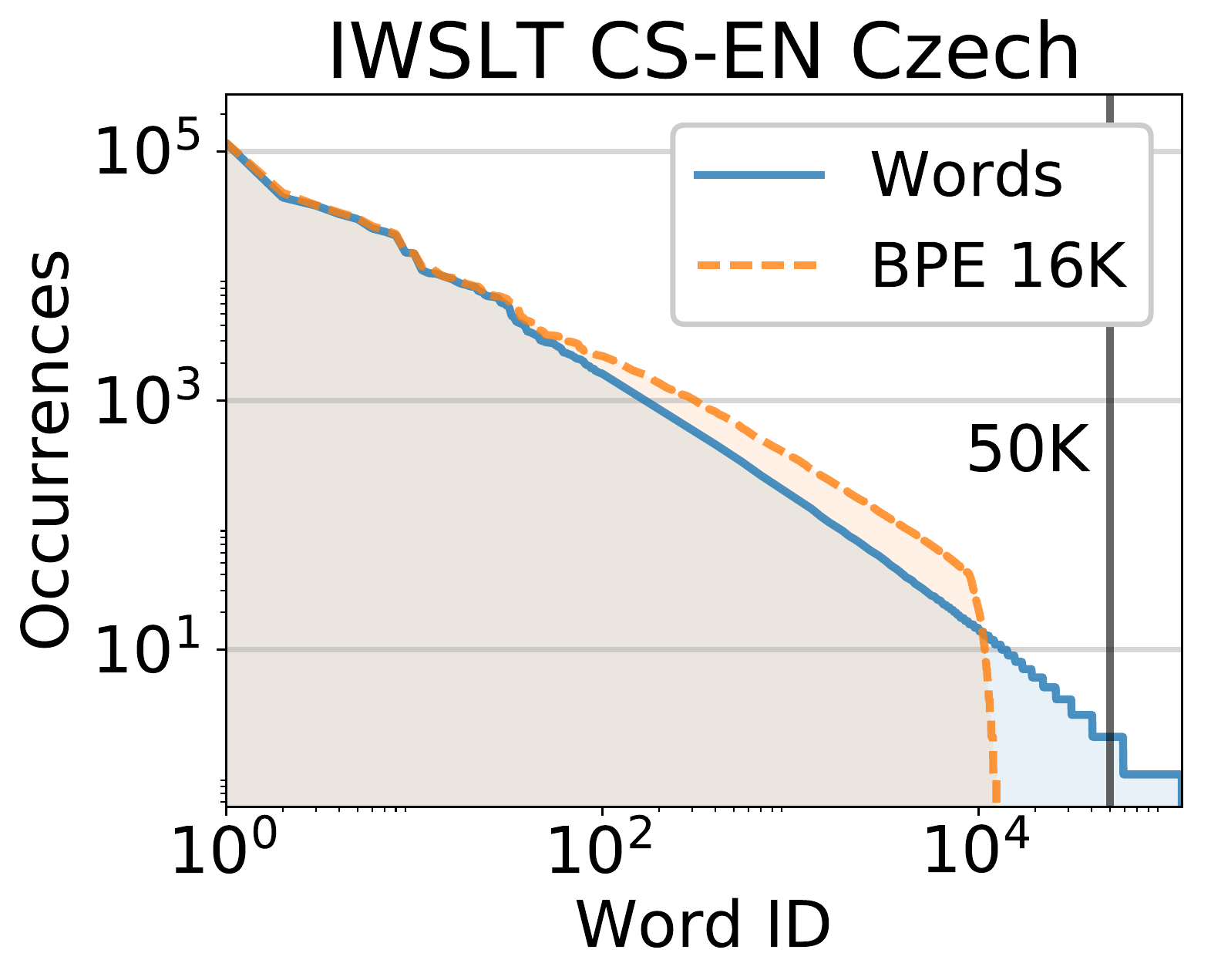}} & \includegraphics[scale=0.325]{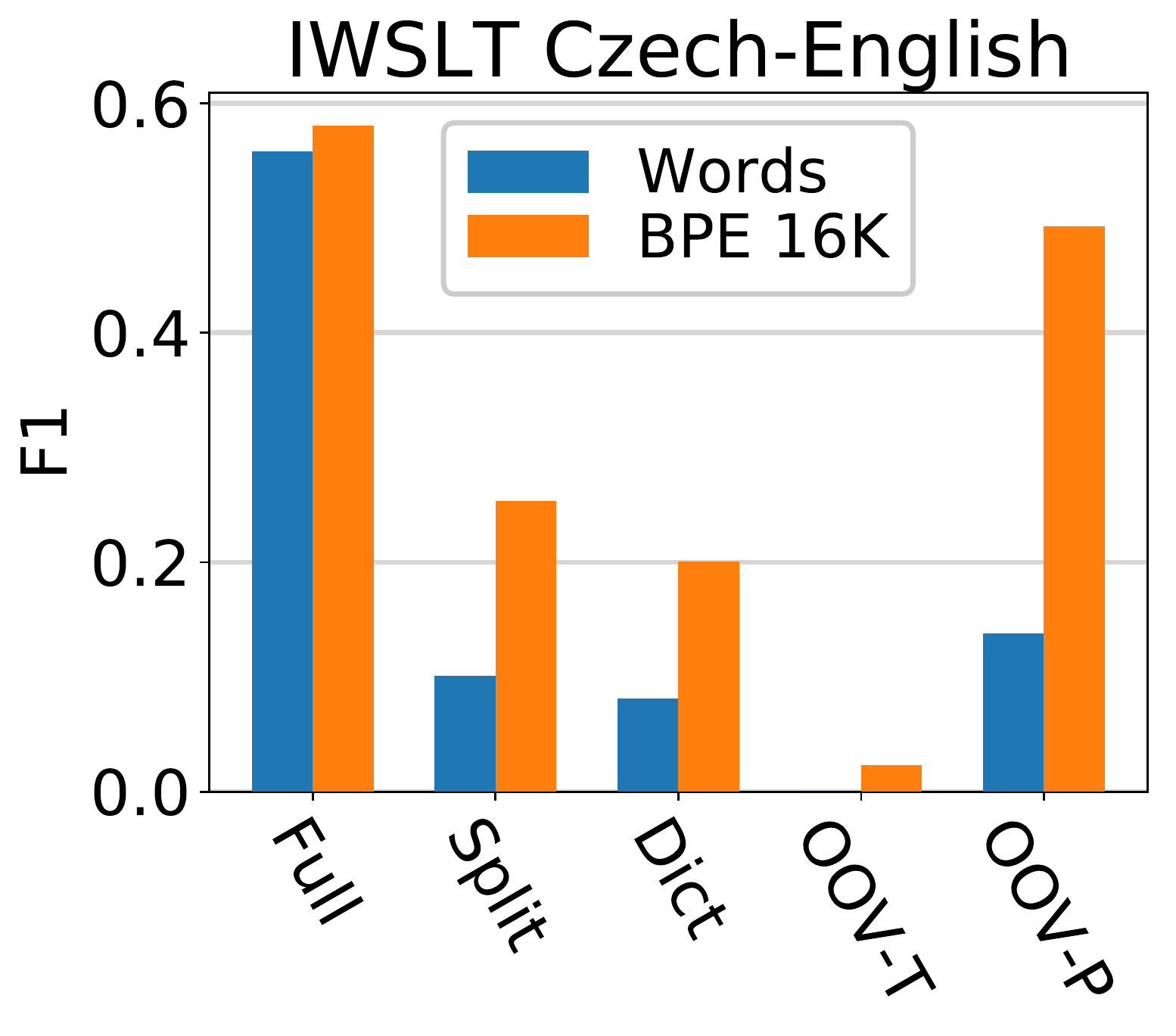} \\
    \end{tabular}
  \end{minipage}\hfill
  \begin{minipage}{0.32\textwidth}
    \caption{\label{fig:bpe}Effects of using sub-word units on model vocabulary and translation accuracy for specific types of words.
      \\\textbf{Left figures:} Source vocabulary visualizations for NMT training data using full words and byte-pair encoded tokens.  The number of merge operations is set to either 32K or 16K, chosen by best BLEU score.  BPE reduces vocabulary size by 1-2 orders of magnitude and allows models to cover the entire training corpus.  Full-word systems for all scenarios use a much larger vocabulary size of 50K (labeled horizontal line) that leaves much of the total vocabulary uncovered.
      \\\textbf{Right figures:} Class-wise test set unigram F1 scores for NMT systems using full words and byte-pair encoded tokens.  Scores are reported separately for the following classes: words in the vocabulary of both the full-word and BPE models (Full), words in the vocabulary of the full-word model that are split in the BPE model (Split), words outside the vocabulary of the full-word model but covered by its dictionary (Dict), words outside the vocabulary of the full-word model and its dictionary that should be translated (OOV-T), and words outside the vocabulary of the full-word model and its dictionary that should be passed through (OOV-P).  All reported scores are averaged over 3 independent optimizer runs.}\vspace{33pt}
  \end{minipage}
\end{figure*}

To understand the origin of these improvements, we divide the words in each test set into classes based on how the full-word and BPE models handle them and report the unigram F-1 score for each model on each class.  We also plot the full-word and BPE vocabularies for context.  As shown in Figure~\ref{fig:bpe}, performance is comparable for the most frequent words that both models represent as single units.  The identical shapes on the leftmost part of each vocabulary plot indicate that the two systems have the same number of training instances from which to learn translations.  For words that are split in the BPE model, performance is tied to data sparsity.  With larger data, performance is comparable as both models have enough training instances to learn reliable statistics; with smaller data or morphologically rich languages such as Finnish, significant gains can be realized by modeling multiple higher-frequency sub-words in place of a single lower-frequency word.  This can be seen as effectively moving to the left in the vocabulary plot where translations are more reliable.  In the next category of words beyond the 50K cutoff, the BPE system's ability to actually model rare words leads to consistent improvement over the full-word system's reliance on dictionary substitution.

The final two categories evaluate handling of true out-of-vocabulary items.  For OOVs that should be translated, the full-word system will always score zero, lacking any mechanism for producing words not in its vocabulary or dictionary.  The more interesting result is in the relatively low scores for OOVs that should simply be copied from source to target.  While phrase-based systems can reliably pass OOVs through 1:1, full-word neural systems must generate \unk{} tokens and correctly map them to source words using attention scores.  Differences in source and target true vocabulary sizes and frequency distributions often lead to different numbers of \unk{} tokens in source and target sentences, resulting in models that are prone to over or under-generating \unk{}s at test time.  BPE systems address these weaknesses, although their performance is not always intuitive.  While some OOVs are successfully translated using word pieces, overall scores are still quite low, indicating only limited success for the notion of open vocabulary translation often associated with sub-word NMT.  However, the ability to learn when to self-translate sub-words\footnote{Learning a single set of BPE operations by concatenating the source and target training data ensures that the same word will always be segmented in the same way whether it appears on the source or target side.} leads to significant gains in pass-through accuracy.

In summary, our analysis indicates that while BPE does lead to smaller, faster models, it also significantly improves translation quality.  Rather than being limited to only rare and unseen words, modeling higher-frequency sub-words in place of lower-frequency full words can lead to significant improvement across the board.  The specific improvement in pass-through OOV handling can be particularly helpful for handling named entities and open-class items such as numbers and URLs without additional dedicated techniques.

\section{Ensembles and Model Diversity}
\label{sec:ensemble}

  The final technique we explore is the combination of multiple translation models into a single, more powerful \textit{ensemble} by averaging their predictions at the word level.  The idea of ensemble averaging is well understood and widely used across machine learning fields and work from the earliest encoder-decoder papers to the most recent system descriptions reports dramatic improvements in BLEU scores for model ensembles \cite{sutskever2014sequence,sennrich-haddow-birch:2016:WMT}.  While this technique is conceptually simple, it requires training and decoding with multiple translation models, often at significant resource costs.  However, these costs are either mitigated or justified when building real-world systems or evaluating techniques that should be applicable to those systems.  Decoding costs can be reduced by using \textit{knowledge distillation} techniques to train a single, compact model to replicate the output of an ensemble \cite{hinton2015distilling,kuncoro-EtAl:2016:EMNLP2016,kim-rush:2016:EMNLP2016}.  Researchers can skip this time-consuming step, evaluating the ensemble directly, while real-world system engineers can rely on it to make deployment of ensembles practical.  To reduce training time, some work ensembles different training checkpoints of the same model rather than using fully independent models \cite{jean-EtAl:2015:ACL-IJCNLP,sennrich-haddow-birch:2016:WMT}.  While checkpoint ensembling is shown to be effective for improving BLEU scores under resource constraints, it does so with less diverse models.  As discussed in recent work and demonstrated in our experiments in \S\ref{sec:eval}, model diversity is a key component in building strong NMT ensembles \cite{jean-EtAl:2015:ACL-IJCNLP,sennrich-haddow-birch:2016:WMT,farajian2016fbk}.  For these reasons, we recommend evaluating new techniques on systems that ensemble multiple independently trained models for the most reliable results.  Results showing both the effectiveness of ensembles and the importance of model diversity are included in the larger experiments conducted in the next section.

\section{On Trustable Evaluation}
\label{sec:eval}

\subsection{Experimental Setup}

\begin{table}
  \small\def\arraystretch{1.2}\setlength\tabcolsep{3pt}
  \begin{center}
    \begin{tabular*}{\linewidth}{@{\extracolsep{\fill}}|l|lll|ll|}
      \hline
      & \multicolumn{3}{c|}{WMT} & \multicolumn{2}{c|}{IWSLT} \\
                                  & DE-EN                & EN-FI                & RO-EN                & EN-FR                & CS-EN                \\
      \hline
      Vanilla & {30.2}               & {11.8}               & {26.4}               & {33.2}               & {20.2}               \\
      \hline
      Recommended & 33.5                 & 14.7                 & 27.8                 & 34.5                 & 22.6                 \\
      +Ensemble & \textbf{35.8} & \textbf{17.3} & \textbf{30.3} & \textbf{37.3} & \textbf{25.5} \\
      \hline
    \end{tabular*}
  \end{center}
  \caption{\label{tab:full}Test set BLEU scores for ``vanilla'' NMT (full words and standard Adam), and our recommended systems (byte pair encoding and annealing Adam, with and without ensembling).  Scores for single models are averaged over 3 independent optimizer runs while scores for ensembles are the result of combining 3 runs.}
\end{table}

In this section, we evaluate and discuss the effects that choice of baseline can have on experimental conclusions regarding neural MT systems.
First, we build systems that include Adam with rate annealing, byte pair encoding, and independent model ensembling and compare them to the vanilla baselines described in \S\ref{sec:system}.
As shown in Table~\ref{tab:full}, combining these techniques leads to a consistent improvement of 4-5 BLEU points across all scenarios.  These improvements are the result of addressing several underlying weaknesses of basic NMT models as described in previous sections, leading to systems that behave much closer to those deployed for real-world tasks.

Next, to empirically demonstrate the importance of evaluating new methods in the context of these stronger systems, we select several techniques shown to improve NMT performance and compare their effects as baseline systems are iteratively strengthened.  Focusing on English-French and Czech-English, we evaluate the following techniques with and without the proposed improvements, reporting results in Table~\ref{tab:experiments}:

\begin{table}
  \small\def\arraystretch{1.2}\setlength\tabcolsep{4pt}
  \begin{center}
    \begin{tabular*}{\linewidth}{@{\extracolsep{\fill}}|l|cc|cc|c|}
      \hline
      EN-FR & \multicolumn{2}{c|}{Adam} & \multicolumn{2}{c|}{+Annealing} & +Ensemble\\
                         & Word      & BPE       & Word      & BPE       & BPE \\
      \hline
      Baseline           & 33.2          & 33.7          & 33.6          & 34.8          & 37.3               \\
      \hline
      Dropout            & \textbf{33.9} & \textbf{33.9} & \textbf{34.5} & \textit{34.7} & \textit{37.2}      \\
      Lexicon Bias       & \textbf{33.8} & \textbf{34.0} & \textbf{33.9} & \textit{34.8} & \textit{37.1}      \\
      Pre-Translation    & --            & \textbf{34.0} & --            & \textbf{34.9} & \textit{36.6}      \\
      Bootstrapping & \textbf{33.7} & \textbf{34.1} & \textbf{34.4} & \textbf{35.2} & \textbf{37.4}      \\
      \hline
      \multicolumn{6}{}{} \\
      \hline
      CS-EN & \multicolumn{2}{c|}{Adam} & \multicolumn{2}{c|}{+Annealing} & +Ensemble\\
                         & Word      & BPE       & Word      & BPE       & BPE \\
      \hline
      Baseline           & 20.2          & 22.1          & 21.0          & 23.0          & 25.5               \\
      \hline
      Dropout            & \textbf{20.7} & \textbf{22.7} & \textbf{21.4} & \textbf{23.6} & \textbf{26.1}      \\
      Lexicon Bias       & \textbf{20.7} & \textbf{22.5} & \textit{20.6} & \textit{22.7} & \textit{25.2}      \\
      Pre-Translation    & --            & \textbf{23.1} & --            & \textbf{23.8} & \textbf{25.8}      \\
      Bootstrapping & \textbf{20.7} & \textbf{23.2} & \textbf{21.6} & \textbf{23.6} & \textbf{26.2}      \\
      \hline
    \end{tabular*}
  \end{center}
  \caption{\label{tab:experiments}Test set BLEU scores for several published NMT extensions.  Entries are evaluated with and without Adam annealing, byte pair encoding, and model ensembling.  A bold score indicates improvement over the baseline while an italic score indicates no change or degradation.  Scores for non-ensembles are averaged over 3 independent optimizer runs and ensembles are the result of combining 3 runs.}
\end{table}

\noindent\textbf{Dropout:} Apply the improved dropout technique for sequence models described by \newcite{NIPS2016_6241} to LSTM layers with a rate of 0.2.  We find this version to significantly outperform standard dropout.

\noindent\textbf{Lexicon bias:} Incorporate scores from a pre-trained lexicon (\texttt{fast\_align} model learned on the same data) directly as additional weights when selecting output words \cite{arthur-neubig-nakamura:2016:EMNLP2016}.  Target word lexicon scores are computed as weighted sums over source words based on attention scores.

\noindent\textbf{Pre-translation:} Translate source sentences with a traditional phrase-based system trained on the same data.  Input for the neural system is the original source sentence concatenated with the PBMT output \cite{DBLP:journals/corr/NiehuesCHW16}.  Input words are prefixed with either \texttt{s\_} or \texttt{t\_} to denote source or target language.  We improve performance with a novel extension where word alignments are used to weave together source and PBMT output so that each original word is immediately followed by its suggested translation from the phrase-based system.  As pre-translation doubles source vocabulary size and input length, we only apply it to sub-word systems to keep complexity reasonable.

\noindent\textbf{Data bootstrapping:} Expand training data by extracting phrase pairs (sub-sentence translation examples) and including them as additional training instances \cite{chen2016guided}.  We apply a novel extension where we train a phrase-based system and use it to re-translate the training data, providing a near-optimal phrase segmentation as a byproduct.  We use these phrases in place of the heuristically chosen phrases in the original work, improving coverage and leading to more fine-grained translation examples.

\subsection{Experimental Results}
\label{sec:experiments}

The immediately noticeable trend from Table~\ref{tab:experiments} is that while all techniques improve basic systems, only a single technique, data bootstrapping, improves the fully strengthened system for both data sets (and barely so).  This can be attributed to a mix of redundancy and incompatibility between the improvements we've discussed in previous sections and the techniques evaluated here.

Lexicon bias and pre-translation both incorporate scores from pre-trained models that are shown to improve handling of rare words.  When NMT models are sub-optimally trained, they can benefit from the suggestions of a better-trained model.  When full-word NMT models struggle to learn translations for infrequent words, they can learn to simply trust the lexical or phrase-based model.  However, when annealing Adam and BPE alleviate these underlying problems, the neural model's accuracy can match or exceed that of the pre-trained model, making external scores either completely redundant or (in the worst case) harmful bias that must be overcome to produce correct translations.  While pre-translation fares better than lexicon bias, it suffers a reversal in one scenario and a significant degradation in the other when moving from a single model to an ensemble.  Even when bias from an external model improves translation, it does so at the cost of diversity by pushing the neural model's preferences toward those of the pre-trained model.  These results further validate claims of the importance of diversity in model ensembles.

Applying dropout significantly improves all configurations of the Czech-English system and some configurations of the English-French system, leveling off with the strongest.  This trend follows previous work showing that dropout combats overfitting of small data, though the point of inflection is worth noting \cite{sennrich-haddow-birch:2016:WMT,DBLP:journals/corr/WuSCLNMKCGMKSJL16}.  Even though the English-French data is still relatively small (220K sentences), BPE leads to a smaller vocabulary of more general translation units, effectively reducing sparsity, while annealing Adam can avoid getting stuck in poor local optima.  These techniques already lead to better generalization without the need for dropout.  Finally, we can observe a few key properties of data bootstrapping, the best performing technique on fully strengthened systems.  Unlike lexicon bias and pre-translation, it modifies only the training data, allowing ``purely neural'' models to be learned from random initialization points.  This preserves model diversity, allowing ensembles to benefit as well as single models.  Further, data bootstrapping is complementary to annealing Adam and BPE; better optimization and a more general vocabulary can make better use of the new training instances.

While evaluation on simple vanilla NMT systems would indicate that all of the techniques in this section lead to significant improvement for both data sets, only evaluation on systems using annealing Adam, byte pair encoding, and independent model ensembling reveals both the reversals of results on state-of-the-art systems and nuanced interactions between techniques that we have reported.  Based on these results, we highly recommend evaluating new techniques on systems that are at least this strong and representative of those deployed for real-world use.

\section{Conclusion}
\label{sec:conclusion}

In this work, we have empirically demonstrated the effectiveness of Adam training with multiple restarts and step size annealing, byte pair encoding, and independent model ensembling both for improving BLEU scores and increasing the reliability of experimental results.  Out of four previously published techniques for improving vanilla NMT, only one, data bootstrapping via phrase extraction, also improves a fully strengthened model across all scenarios.  For these reasons, we recommend evaluating new model extensions and algorithms on NMT systems at least as strong as those we have described for maximally trustable results.

\bibliographystyle{acl_natbib}
\bibliography{refs}

\end{document}